\documentclass[12pt]{article}

\usepackage{fontspec}
\newfontfamily{\arabicfont}[Path=./,Script=Arabic]{DejaVuSans.ttf}
\newcommand{\ara}[1]{{\arabicfont #1}}   

\usepackage[a4paper, margin=2.5cm]{geometry}

\usepackage{amsmath}
\usepackage{amssymb}

\usepackage{booktabs}
\usepackage{graphicx}
\usepackage{xcolor}

\usepackage{hyperref}
\hypersetup{
  colorlinks=true,
  linkcolor=black,
  citecolor=black,
  urlcolor=blue,
  pdftitle={Quantum Compositional NLP for Arabic: Grammar, Morphology, and Word Sense in Circuit Topology},
  pdfauthor={Wajahath Mohammed},
  pdfkeywords={quantum NLP, Arabic NLP, DisCoCat, pregroup grammar, word order, word sense disambiguation, lambeq},
  pdfsubject={Quantum Natural Language Processing}
}

\usepackage{microtype}
\usepackage{setspace}
\usepackage{footmisc}

\title{\textbf{Quantum Compositional NLP for Arabic:\\
Grammar, Morphology, and Word Sense in Circuit Topology}}

\author{Wajahath Mohammed\\
\texttt{wajahath123@gmail.com}\\[0.4em]
\small Preprint: \url{https://doi.org/10.5281/zenodo.19564468}}

\date{}

\begin{document}

\maketitle

\begin{abstract}
We present the first application of pregroup grammar-based quantum compositional natural
language processing (QNLP) to Arabic, a morphologically rich, free-word-order language
whose structural complexity provides a uniquely demanding testbed for theories of meaning
composition in quantum circuits. Our system converts Arabic sentences into quantum circuits
whose topology mirrors grammatical structure: subjects, verbs, and objects become quantum
gates, and the typed dependencies between them (the pregroup grammar) determine how
those gates are wired together. We conduct three controlled experiments spanning word order,
morphological tense, and verb sense disambiguation, comparing quantum circuit methods against
classical baselines including AraVec (Arabic word embeddings) and AraBERT (a pre-trained
Arabic transformer). The central finding is a clean causal ablation: quantum circuits encoding
only grammar topology, with no parameterised components, achieve exactly 50\% on a matched-pair
word order task where classical bag-of-words models fail (12.8\%); adding a single layer of
parameterised entangling gates raises performance to 64.9\% (mean $\pm$ 3.2\% std across 10
seeds, 15 folds each). This 15-percentage-point gain (with zero variance at the ablation
baseline across all seeds and folds) is entirely attributable to parameterised entanglement,
establishing a clean causal claim. We additionally introduce the first vocabulary-controlled Arabic word sense
disambiguation dataset, using matched sentence pairs and shared lexical pools to isolate
structural from lexical disambiguation signals, and characterise a SPSA label inversion
phenomenon whose rate is measurably reduced by ancilla qubit encoding. Arabic's Semitic
root-and-pattern morphology has a formal correspondence to quantum tensor products,
identified independently in the computational linguistics literature, which no other language
in the existing QNLP literature shares, positioning Arabic as a theoretically motivated and
practically significant target for quantum compositional methods.
\end{abstract}

\tableofcontents
\newpage

\section{Introduction}

When a sentence is processed by a computer, one of the most consequential decisions a system
makes is what to throw away. Classical language models built on word embeddings (the
dominant paradigm from the mid-2010s through the rise of transformers) discard word order
almost entirely. A sentence is treated as a bag: its words are averaged, summed, or
concatenated, and the resulting representation carries no information about whether
\textit{the student read the book} or \textit{the book read the student}. This approximation
works surprisingly well across many tasks because, in most languages, the most common word
order is consistent enough that vocabulary alone is a strong proxy for meaning.

Arabic is one of the languages where this approximation breaks most severely, and for
principled reasons that connect directly to the mathematical structure of quantum computation.

Arabic is the native language of over 400 million people, one of six official United Nations
languages, and the liturgical language of 1.8 billion Muslims worldwide. It possesses a
morphological system of documented complexity: every word is derived from a consonantal root
(typically three letters) combined with a vowel pattern that encodes grammatical properties:
aspect, voice, derived stem, number, gender. The verb \textit{kataba} (he wrote),
\textit{yaktubu} (he writes), \textit{katib} (writer), \textit{maktub} (written), and
\textit{kitab} (book) all derive from the single root k-t-b; they share no surface phonological
material with each other beyond those three consonants, yet they are systematically related by
the abstract root. Word order is free by typological standards: Arabic sentences occur in
Subject-Verb-Object (SVO), Verb-Subject-Object (VSO), and Nominal (subjectless predication)
forms in the same written register, with the choice determined by pragmatic focus rather than
grammatical constraint.

These properties make Arabic an ideal testbed for quantum compositional NLP (QNLP). The
DisCoCat framework \cite{coecke2010} maps the type-theoretic structure of pregroup grammar
onto quantum circuits: words become qubit states, grammatical dependencies become entangling
gates, and the sentence-level meaning is a measurement outcome whose probability depends on
how the word-states interacted through the grammar. Different word orders produce topologically
different circuits. Morphologically distinct verb forms produce circuits with different internal
structure. The claim of the quantum compositional framework is that this structural encoding is
not merely a notation; it carries genuine representational content that is measurable and
distinct from what classical order-blind models can access.

Kiraz's (2001) formal analysis of Semitic morphology provides an additional, non-obvious
connection. He proved that Arabic root-and-pattern morphology requires multi-tape finite
automata, computations that proceed on multiple parallel information streams simultaneously.
The consonantal root runs on one tape; the vowel pattern encoding grammatical categories runs
on another; meaning emerges from their interaction. Tensor products in quantum mechanics
formalise exactly this notion of parallel composition: the joint state of two independent
systems is their tensor product. This is not an arbitrary analogy: parallel composition of
simultaneous information streams is the structural problem that both multi-tape automata and
tensor products formalise. The connection between Kiraz's result and quantum tensor products is
a motivated interpretive step made in the present work, stated explicitly as such in
Section~\ref{sec:discussion_arabic}. No other language in the existing QNLP literature has a
morphological structure with this formal parallel-composition property.

This paper asks whether these theoretical connections are genuine and empirically measurable.
We approach the question through three controlled experiments, each designed to test a specific
structural property. The experiments are not designed to show that quantum models outperform
state-of-the-art classical systems; AraBERT fine-tuned on even small datasets achieves
near-perfect accuracy on all tasks. The experiments are designed to demonstrate that quantum
circuits encode structural information through a \emph{mechanistically distinct} pathway,
one that is provably unavailable to order-blind models, causally attributable to entanglement
through a zero-variance ablation, and formally connected to properties unique to Arabic among
all languages currently in the QNLP literature.

\noindent\textbf{Contributions:}
\begin{enumerate}
  \item The first pregroup grammar-based QNLP system for Arabic, where circuit topology is
        derived from grammatical structure rather than classical language model embeddings,
        including Arabic-specific pregroup grammar rules for SVO, VSO, and Nominal sentence
        structures.
  \item A controlled ablation demonstrating that parameterised entanglement in IQP circuits
        accounts for a 15\,pp gain on matched-pair Arabic word order classification (QFM L0:
        50.0\% exactly; QFM L1: 64.9\%).
  \item The first vocabulary-controlled Arabic word sense disambiguation dataset, comprising
        200 sentences across four verbs with matched pairs, shared object pools, and shared
        subject pools designed to isolate structural from lexical disambiguation signals.
  \item Characterisation of SPSA label inversion in symmetric binary tasks, including a
        measurement of ancilla qubit encoding's effect on inversion rate (99\% $\to$ 23\% on
        the hardest task).
\end{enumerate}

\section{Background}

\subsection{Arabic Linguistic Structure}

The two structural properties most relevant to the present work are the root-and-pattern
morphological system and the free word order.

\textbf{Root-and-pattern morphology.} In Arabic, the consonantal root is an abstract morpheme
with no pronunciation of its own. It combines with a vowel-and-consonant pattern to produce a
surface form that is simultaneously a phonological word and a grammatical category. The pattern
\textit{CaCaCa} (where C marks consonant slots) produces third-person masculine singular
past-tense verbs; the pattern \textit{ya-CC-uCu} produces third-person masculine singular
present-tense verbs. The meaning of the word form is the interaction of root meaning and
pattern meaning. This is compositional in the technical sense: the meaning of the whole is a
function of the meanings of the parts and how they combine. What makes it unusual is the
\textit{parallel} nature of the combination: root and pattern are not concatenated sequentially
but interleaved simultaneously. Kiraz (2001) formalised this as multi-tape computation. Quantum
tensor products formalise parallel composition of independent registers. Both formalisms
describe the same structural property (meaning as the interaction of simultaneously-processed
information streams), applied to the same morphological problem. This structural alignment
across formalisms is the theoretical anchor of the future work described in
Section~\ref{sec:future_morphological}; it is stated explicitly as an interpretive step in
Section~\ref{sec:discussion_arabic}, not as a direct theorem of Kiraz's analysis.

\textbf{Free word order.} Arabic allows three major word order types as fully grammatical
alternatives within Modern Standard Arabic: SVO (\textit{al-walad kataba al-dars}, ``the boy
wrote the lesson''), VSO (\textit{kataba al-walad al-dars}, ``wrote the boy the lesson''), and
Nominal sentences without a verb (\textit{al-walad najib}, ``the boy is intelligent''). SVO
order topicalises the subject; VSO order topicalises the action. Both are used in formal
written Arabic. This three-way alternation, where the same three content words can appear in
different orders with different grammatical structures, creates a natural testbed for models
that claim to encode structural information: a model must produce different representations for
the same words in different orders if it is to classify word order correctly.

\subsection{DisCoCat and Pregroup Grammar}

DisCoCat \cite{coecke2010} is a compositional theory of meaning grounded in category theory.
Each word is assigned a grammatical type from a pregroup: a noun has type $n$, a transitive
verb has type $n^r \otimes s \otimes n^l$ (a function cancelling a noun on its left and a noun
on its right to produce a sentence type $s$). A sentence is grammatical if and only if the
tensor product of its word types reduces to $s$ via the pregroup reduction rules (cup
operations). The string diagram for this reduction is a planar graph whose topology encodes the
grammatical structure.

In the quantum setting, this string diagram is compiled into a quantum circuit: word types
become qubit registers, cups become entangling gates (specifically, post-selection or partial
traces), and the sentence-level meaning is a measurement of the sentence qubits. The key
property for this paper is that \emph{the circuit topology is uniquely determined by the
grammatical analysis}: sentences with different grammatical structures produce circuits with
different topologies, regardless of vocabulary.

\subsection{IQP Circuits and the Ablation Design}

The Instantaneous Quantum Polynomial (IQP) ansatz constructs circuits where noun-type words
are represented as single qubits initialised by parameterised Hadamard-and-Z-rotation gates,
and entanglement between qubits is introduced by parameterised controlled-Z rotation gates.
The depth of entanglement (the number of entangling layers) is a hyperparameter. At depth
zero (L0), no entangling gates are applied; each qubit evolves independently. At depth one (L1)
or two (L2), entangling gates couple adjacent qubits according to the diagram topology.

The L0/L1 comparison is the paper's central experimental design. In QFM mode, circuit
parameters are fixed to values derived from word embeddings. At depth zero (L0), there are no
entangling gates: word qubits evolve independently, and no information from word qubits can
propagate to the sentence qubit. The sentence qubit does not correspond to any word and has no
embedding-derived parameter; it receives no gates at L0 and thus produces a constant output,
predicting the same class for every input sentence regardless of its structure or vocabulary. On a balanced binary dataset, constant prediction gives exactly 50\% accuracy.
Crucially, it does so with \emph{zero variance}: because the output is deterministic (not
random), it does not fluctuate across seeds or cross-validation folds. This is not a
measurement; it is a structural property of the circuit architecture. The depth-zero circuit
therefore serves as a zero-variance baseline: any improvement at depth L1 is caused entirely
by the entangling gates that allow word qubit states to influence the sentence qubit's output.

\section{Related Work}

\subsection{QNLP: Foundations and Hardware}

The theoretical foundations of the present work are the original DisCoCat paper
\cite{coecke2010} and the near-term hardware framework of Meichanetzidis et al.\ (2020)
\cite{meichanetzidis2020}. Meichanetzidis et al.\ provided the first full-stack description of
how DisCoCat string diagrams map to NISQ-compatible quantum circuits, establishing the pipeline
that later became the lambeq library \cite{kartsaklis2021}. The lambeq library is the primary
implementation framework for this work.

The first empirical QNLP experiments at scale were reported by Lorenz et al.\ \cite{lorenz2023},
who classified English sentences into food and IT topics on real IBM quantum hardware and
demonstrated that the syntax-sensitive DisCoCat model is trainable on device. Their dataset
size (approximately 130 sentences) is comparable to the datasets used here. The most recent
hardware milestone is Duneau et al.\ \cite{duneau2024}, who ran the DisCoCirc text-level model
on Quantinuum's H1-1 trapped-ion processor and demonstrated compositional generalisation ---
the ability to correctly process sentence combinations that were not present in training ---
on a task where neither GPT-4, LSTMs, nor transformer baselines succeeded. This result is
particularly relevant to the present work: it establishes that quantum compositional models
have a measurable advantage on compositional generalisation tasks at the level of hardware
execution, and that this advantage is not replicated by contemporary large language models.

The methodological design of the present work is directly validated by a 2025 survey
\cite{nausheen2025}, which identifies systematic entanglement layer ablation as an outstanding
open problem across the reviewed literature and lists Arabic as a notable gap in QNLP language
coverage. The L0/L1 ablation of Experiment 1 is a direct response to the first gap; the
language choice addresses the second.

\subsection{QNLP Beyond English}

The field has begun extending beyond English in the past three years. Three bodies of work
are directly related.

\textbf{Urdu and DisCoCirc \cite{waseem2022}.} Waseem, Liu, Wang-Ma\'scianica, and Coecke
applied the DisCoCirc framework to Urdu, demonstrating that the circuit-level representations
of English and Urdu sentences converge despite their surface differences in word order. Their
paper is a theoretical contribution: it shows that the DisCoCirc framework is in principle
language-independent. It does not contain classification experiments, training results, or
vocabulary-controlled evaluation.

\textbf{Persian QNLP \cite{abbaszade2023}.} Abbaszade and Zomorodi applied DisCoCat to
machine translation between English and Persian, achieving low mean absolute error on a
160-sentence corpus. Persian uses Arabic script and has borrowed extensively from Arabic
vocabulary, but its grammar is structurally distinct from Arabic in the ways that matter here:
Persian morphology is agglutinative rather than root-pattern, and Persian word order is
predominantly SOV rather than the three-way alternation of Arabic. The formal Kiraz
correspondence does not apply to Persian. Sadrzadeh's earlier pregroup grammar analysis of
Persian sentences \cite{sadrzadeh2007} provides the formal linguistic foundation that
Abbaszade and Zomorodi reference.

\textbf{Hindi QNLP \cite{srivastava2023}.} Srivastava, Babu, and colleagues constructed
pregroup grammar derivations for Hindi within the DisCoCat framework and trained
grammar-aware and topic-aware sentence classifiers using lambeq and IQP circuits. Their paper
is the most similar in form to the present work (classification experiments, lambeq, IQP
circuits) but does not employ vocabulary-controlled evaluation, does not report an L0/L1
ablation, and does not address the word order classification problem specifically.

\textbf{Arabic Hybrid Quantum-Classical Classification \cite{djemmal2025}.} Djemmal and
Belhadef published AraBERT-QC in the Journal of Supercomputing (2025): AraBERT's [CLS]
embeddings are dimensionality-reduced via PCA and Haar transform, then fed into a
parameterised quantum circuit (PQC) for short Arabic sentence classification. This is the
closest contemporary work in surface appearance. The key structural distinction is that
AraBERT-QC uses classical contextual embeddings as the input to its quantum component; the
quantum circuit processes AraBERT's representation, not a pregroup grammar derivation. The
circuit topology does not encode grammatical structure; it is a generic trainable layer applied
to dense vectors. There is no entanglement ablation, no vocabulary-controlled evaluation, no
pregroup type assignment, and no structural inductive bias analysis. The present work constructs
circuits whose topology is entirely determined by the pregroup grammar derivation of each
sentence, with no classical language model in the pipeline, and isolates the causal
contribution of entanglement through a zero-variance L0 baseline.

\subsection{Arabic in Formal and Computational Linguistics}

Arabic has an extensive history of formal grammatical analysis. The classical Arabic grammatical
tradition (\textit{nahw}), dating from S\={\i}bawayhi's eighth-century \textit{Al-Kit\=ab},
developed type-theoretic analyses of Arabic syntax that are strikingly compatible with
pregroup grammar. Sadrzadeh \cite{sadrzadeh2007}, one of the three DisCoCat founders,
applied pregroup grammar analysis to Persian sentences, establishing the precedent for extending
pregroup analysis beyond Indo-European languages.

On the computational side, Habash \cite{habash2010} identifies morphological disambiguation as
``the central hard problem of Arabic NLP,'' a bottleneck that limits performance on
downstream tasks including information retrieval, machine translation, and speech recognition.
Current systems use statistical disambiguation models trained on millions of manually annotated
tokens \cite{pasha2014}. A quantum model with explicit root and pattern qubit registers could
approach this problem from the compositional side, with generalisation capacity that comes from
the mathematical structure rather than training data alone.

\subsection{Vocabulary-Controlled Evaluation}

The matched-pair evaluation methodology introduced in this paper draws on a tradition of
minimal-pair testing in linguistics and NLP. BLiMP \cite{warstadt2020} constructs 67 datasets
of 1,000 minimal sentence pairs for English, covering morphology, syntax, and semantics, with
pairs that are nearly identical in vocabulary but differ in a structural property. Our
matched-pair word order construction extends this logic in a specific direction: the same
lexical items appear with different structural labels in the classification training data,
forcing the model to use structural information rather than lexical co-occurrence statistics.
To the best of our knowledge, no prior QNLP paper has employed a vocabulary-controlled
evaluation design of this kind, and no prior Arabic NLP paper has applied minimal-pair
evaluation to word order classification.

\section{The Arabic QNLP Pipeline}

\subsection{The Problem of Connecting Arabic to Quantum Circuits}

The path from an Arabic sentence to a quantum circuit is not straightforward, and the
development of a robust pipeline was a substantial technical undertaking that preceded the
experiments described in this paper.

Existing QNLP frameworks were designed around English. English has fixed
subject-verb-object order, simple morphology, and extensive NLP tooling. Arabic has three
valid word orders, rich morphology requiring specialised analysis tools, and is written
right-to-left in a script that is systematically different from the Latin alphabet assumed by
standard NLP parsers. The lambeq library, at the time of this work, had no native handling for
Arabic, no VSO grammar rules, and no integration with Arabic morphological analysers.

Early development work addressed the Arabic analysis problem. CAMeL Tools \cite{obeid2020},
the standard Arabic NLP library, was used for morphological analysis: given an Arabic
word form, it returns the root, pattern, aspect, person, number, gender, definiteness, and
case. Stanza \cite{qi2020}, a general-purpose dependency parser, was used for syntactic
structure: it identifies the head verb, the subject noun, and the object noun by their
dependency relations. A recurring problem was tokenisation disagreement between CAMeL Tools
and Stanza: cliticised words (where a conjunction or preposition is attached to the following
word) are segmented differently by the two tools. A fallback chain handles these cases by
attempting multiple analysis strategies and defaulting to a minimum-valid diagram when all
else fails.

The central unresolved problem in early experiments was type consistency: lambeq's diagram
system requires that every diagram reduce to the sentence type $s$, meaning the grammatical
types must ``cancel'' correctly. For SVO sentences, the reduction follows the standard English
pattern. For VSO sentences, where the verb appears before both its subject and object,
the type of the verb must be formulated as a function that looks forward for both arguments,
requiring a different type formula ($s \otimes n^l \otimes n^l$ rather than
$n^r \otimes s \otimes n^l$) and an additional Swap operation in the diagram. This was not
handled by any existing lambeq parser. The \texttt{arabic\_dep\_reader.py} module implements
this from scratch.

\subsection{Pipeline Architecture}
\label{sec:pipeline}

The final pipeline consists of five stages.

\textbf{Stage 1: Morphological Analysis.} Each sentence is parsed by Stanza for dependency
structure and simultaneously analysed by CAMeL Tools for morphological features. Features are
encoded as string tags appended to the word form: \textit{kataba} becomes
\texttt{كتب\_ASP-p\_PER-3\_NUM-s\_GEN-m} (past aspect, third person, singular, masculine).
These morphologically enriched labels are used as word identities throughout the diagram and
circuit construction, ensuring that distinct morphological forms produce distinct circuit
parameters.

\textbf{Stage 2: Structure Detection.} Based on the relative positions of the identified
subject, verb, and object, the sentence is classified as SVO, VSO, SV (intransitive,
subject-first), VS (intransitive, verb-first), Nominal, or Fallback. A verb-rescue mechanism
handles Stanza's known failure mode on Arabic light verb constructions, falling back to the
first VERB-tagged token in the parse if the dependency root is not identified as a verb.

\textbf{Stage 3: Diagram Construction.} \texttt{arabic\_dep\_reader.py} constructs a
pregroup grammar diagram. For SVO:
\[
  n \otimes (n^r \otimes s \otimes n^l) \otimes n \;\longrightarrow\; s,
  \quad\text{contracted by } \mathrm{Cup}(n,n^r) \otimes \mathrm{Id}(s) \otimes \mathrm{Cup}(n^l,n)
\]
For VSO, a Swap operation is required to align the two backward-pointing types with their noun
arguments:
\[
  (s \otimes n^l \otimes n^l) \otimes n \otimes n \;\longrightarrow\; s,
  \quad\text{after } \mathrm{Swap}(n^l, n)\text{ at position 2--3, contracted by two cups}
\]

The VSO diagram has a structurally distinct topology from the SVO diagram: the additional Swap
operation in the VSO case produces a different gate count after compilation, specifically fewer
controlled gates in the compiled IQP circuit. This structural difference is the signature that
the topology-only classifier exploits.

\textbf{Why the Swap is not an engineering choice.} The Swap is not an ad hoc workaround;
it is the unique minimal operation required by the pregroup type mathematics. When a VSO verb
is assigned type $s \otimes n^l \otimes n^l$, the tensored diagram has wire sequence
$s,\, n^l,\, n^l,\, n,\, n$. To fire the cups that reduce this to $s$, the second $n^l$ must
cross the first $n$; there is no other valid reduction path. The Swap is the proof that the
verb type was assigned correctly. This is what distinguishes the pipeline from approaches that
label a sentence as ``VSO'' without changing the circuit topology: the label alone does
nothing; the different verb type, and the Swap it necessitates, are what produce structurally
distinct circuits for SVO and VSO sentences containing identical words.

\textbf{Concrete walkthrough.} Consider the matched pair used in Experiment 1. The SVO
sentence \ara{الولد كتب الدرسَ} (\textit{al-waladu kataba al-darsa}, meaning ``The boy wrote the
lesson'') is analysed by Stanza as: \ara{الولد} = nsubj, \ara{كتب} = root (verb), \ara{الدرس} = obj.
Since the subject precedes the verb, the verb is assigned type $n^r \otimes s \otimes n^l$,
and the reduction proceeds as:
\[
  n\;(\text{boy}) \otimes (n^r \otimes s \otimes n^l)\;(\text{wrote}) \otimes n\;(\text{lesson}) \to s
\]
Now take the VSO counterpart \ara{كتب الولدُ الدرسَ} (\textit{kataba al-waladu al-darsa},
the same words in verb-first order). The verb now precedes both arguments, so it is assigned type
$s \otimes n^l \otimes n^l$, requiring a Swap before the cups can fire. The circuit compiled
from the VSO diagram at IQP depth L1 has a different controlled-gate arrangement than the SVO
circuit, not because the words are different (they are identical), but because the grammar
produces a different string diagram. This is the difference the classifier detects.

\textbf{Stage 4: Circuit Compilation.} lambeq's \texttt{RemoveCupsRewriter} simplifies the
diagram by removing cup operations, and \texttt{IQPAnsatz} converts the result into a quantum
circuit. At depth L0, no entangling gates are applied. At L1 or L2, controlled-Z rotation
gates are introduced between adjacent qubits. The \texttt{NumpyModel} backend simulates the
circuit.

\textbf{Stage 5: Classification.} Two strategies: (a) Quantum Feature Map (QFM), whereby
the circuit with parameters fixed to word embedding values serves as a fixed feature extractor
and a classical SVM classifies the resulting output vectors (10 seeds, results averaged); (b)
SPSA training, in which circuit parameters are optimised end-to-end with the SPSA optimiser over
300--500 epochs (5 seeds $\times$ 15 folds).

\subsection{Technical Challenges}

Three engineering challenges deserve documentation as both record and caution for future work.

\textbf{NumPy version incompatibility.} lambeq 0.5.0 was compiled against NumPy 1.x. All
experiments use a dedicated virtual environment with NumPy 1.x, isolated from the system
Python which had NumPy 2.x.

\textbf{Ancilla density matrix encoding.} \texttt{IQPAnsatz} with \texttt{discard=True,
n\_ancillas=1} produces NumpyModel output of shape \texttt{(batch, 2, 2)}, density matrices
for a one-qubit traced-out system, rather than \texttt{(batch, 2)} ket vectors. The standard
one-hot label encoding is incompatible with this output shape. The fix required a custom label
encoding producing pure-state density matrices (class 0 $\to [[1,0],[0,0]]$, class 1 $\to
[[0,0],[0,1]]$) and a custom prediction function extracting the diagonal.

\textbf{SPSA label inversion.} Documented in Section~\ref{sec:inversion}. SPSA can converge
to the inverted solution (correct separation, wrong orientation). This is a known phenomenon
in binary optimisation but has not been previously characterised in the QNLP literature.

\section{Datasets and Experimental Design}

\subsection{The Sentence Corpus}

All experiments draw from a JSON corpus (\texttt{sentences.json}) of 1,140 sentences stored
across seven dataset keys, constructed manually in Modern Standard Arabic and annotated with
class labels.\footnote{\texttt{TenseBinary} (100 sentences) is a cleaner binary re-extraction
of the tense subset already present in \texttt{Morphology}; it shares sentences with that key
and was generated by \texttt{generate\_exp13\_data.py}. \texttt{WordSenseDisambiguation}
(160 sentences, v1) was superseded by the vocabulary-controlled
\texttt{WordSenseDisambiguation\_v2} (200 sentences) and is not used in any experiment reported
in this paper. The total of uniquely authored sentences is therefore 1,040 (1,140 $-$ 100
TenseBinary overlap), but all seven keys are retained in the corpus file for provenance.}

\begin{table}[h]
\centering
\caption{Sentence corpus dataset keys.}
\label{tab:corpus}
\begin{tabular}{llll}
\toprule
Dataset Key & $N$ & Classes & Structure \\
\midrule
WordOrder            & 120 & SVO / VSO / Nominal  & 40/class \\
WordOrderMatched     & 120 & SVO / VSO            & 60/class, matched pairs \\
LexicalAmbiguity     & 210 & 14 sense classes     & 15/class \\
Morphology           & 230 & Tense, Number, Poss. & Mixed \\
TenseBinary          & 100 & Past / Present       & 50/class \\
WordSenseDisambiguation   & 160 & 8 verb-sense classes & 20/class \\
WordSenseDisambiguation\_v2 & 200 & 8 verb-sense classes & 25/class \\
\bottomrule
\end{tabular}
\end{table}

The \texttt{WordOrderMatched} key is the core dataset for Experiment 1. Every sentence is a
member of a matched pair: the same three words appear in SVO and VSO order with different
labels. A model that represents sentences as sums or averages of word vectors produces
identical representations for both members, and therefore cannot exceed 50\% on this dataset by any
mechanism that uses only lexical identity.

\subsection{Methods}
\label{sec:methods}

\textbf{AraVec (Bag-of-Words).} Mean word vectors from the 300-dimensional AraVec Twitter
model \cite{soliman2017}. SVM with RBF kernel. This is the order-blind lexical baseline; its
performance on matched-pair tasks provides a controlled lower bound.

\textbf{AraBERT (Frozen CLS).} [CLS] token from \texttt{aubmindlab/bert-base-arabertv02}
\cite{antoun2020}, no fine-tuning. Represents contextual representations without task-specific
adaptation.

\textbf{AraBERT (Fine-tuned).} AraBERT with classification head, fine-tuned for 10 epochs per
fold using a manual PyTorch training loop. Represents the practical upper bound of contemporary
supervised NLP.\footnote{Fine-tuned AraBERT was run in a dedicated script isolated from the
quantum circuit experiments, using a manual PyTorch training loop (the HuggingFace Trainer API
was incompatible with the installed version of \texttt{accelerate}). Results are stored
separately in \texttt{arabert\_finetuned\_results.json}.}

\textbf{Topology-only.} Counts entangling (Controlled) gates in the compiled circuit as the
sole feature. Zero parameters. Tests whether the parser alone, without any learning, separates
classes structurally.

\textbf{QFM (Quantum Feature Map).} Circuits evaluated with random parameters as fixed feature
extractors for an SVM. No quantum training. Averaged over 10 seeds. Isolates the
representational capacity of circuit structure from parameter optimisation.

\textbf{SPSA (Quantum Training).} Full end-to-end quantum circuit training. 300--500 epochs,
batch size 8, IQP ansatz. 5 seeds $\times$ 15 folds.

Cross-validation: 5-fold stratified, 3 repeats (15 splits). 95\% normal approximation
confidence intervals for QFM, computed across all 150 fold-seed evaluations (10 random seeds
$\times$ 15 stratified cross-validation folds).\footnote{Confidence intervals computed as
mean $\pm$ 1.96 $\times$ (std/$\sqrt{n}$) across $n = 150$ evaluations (10 random seeds
$\times$ 15 stratified cross-validation folds), treating each fold-seed pair as an independent
evaluation. This is a normal approximation; cross-validation folds within a seed are
correlated, so intervals are indicative rather than exact.} Confidence intervals are reported
for Experiment 1 (word order) only; the tense and WSD experiments employed different
evaluation protocols and CIs for those results are not reported here, a limitation acknowledged
in Section~\ref{sec:limitations}.

\section{Experiment 1: Word Order and the Core Ablation}

\subsection{Design}

Binary classification (SVO vs.\ VSO) on the 120 matched-pair sentences of
\texttt{WordOrderMatched}. Vocabulary is fully controlled. Any model exceeding 50\% uses
information beyond lexical identity.

\subsection{Results}

\begin{figure}[t]
  \centering
  \includegraphics[width=0.80\textwidth]{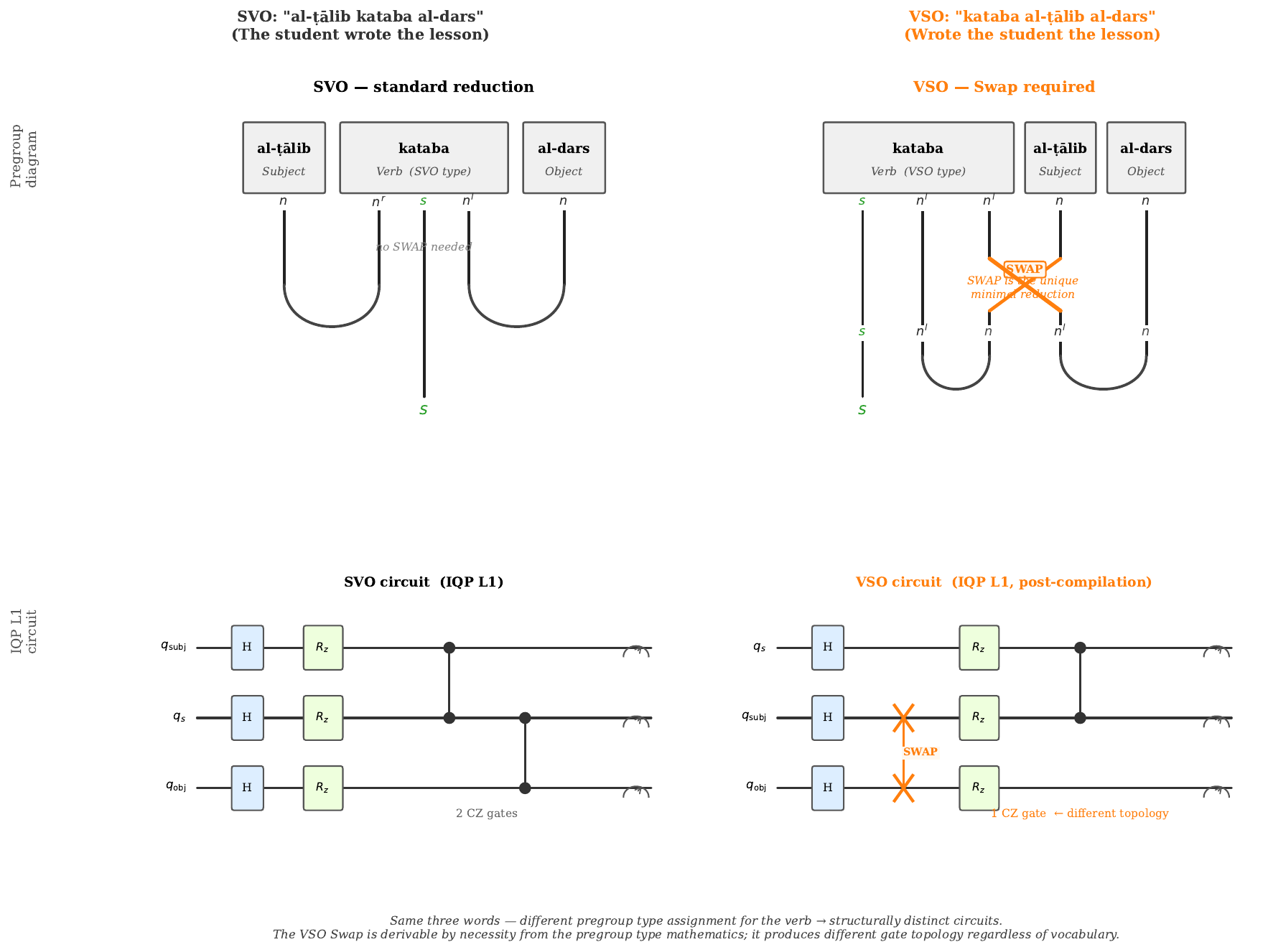}
  \caption{SVO (left) and VSO (right) pregroup string diagrams with compiled IQP L1 circuits
    below. SVO reduces via two cups with no Swap. VSO requires a Swap (highlighted in orange)
    before the cups can fire: the unique minimal operation derivable from the VSO verb type
    $s \otimes n^l \otimes n^l$. The compiled VSO circuit has fewer CZ gates, producing
    structurally distinct topology regardless of vocabulary.}
  \label{fig:diagrams}
\end{figure}

\begin{figure}[t]
  \centering
  \includegraphics[width=0.75\textwidth]{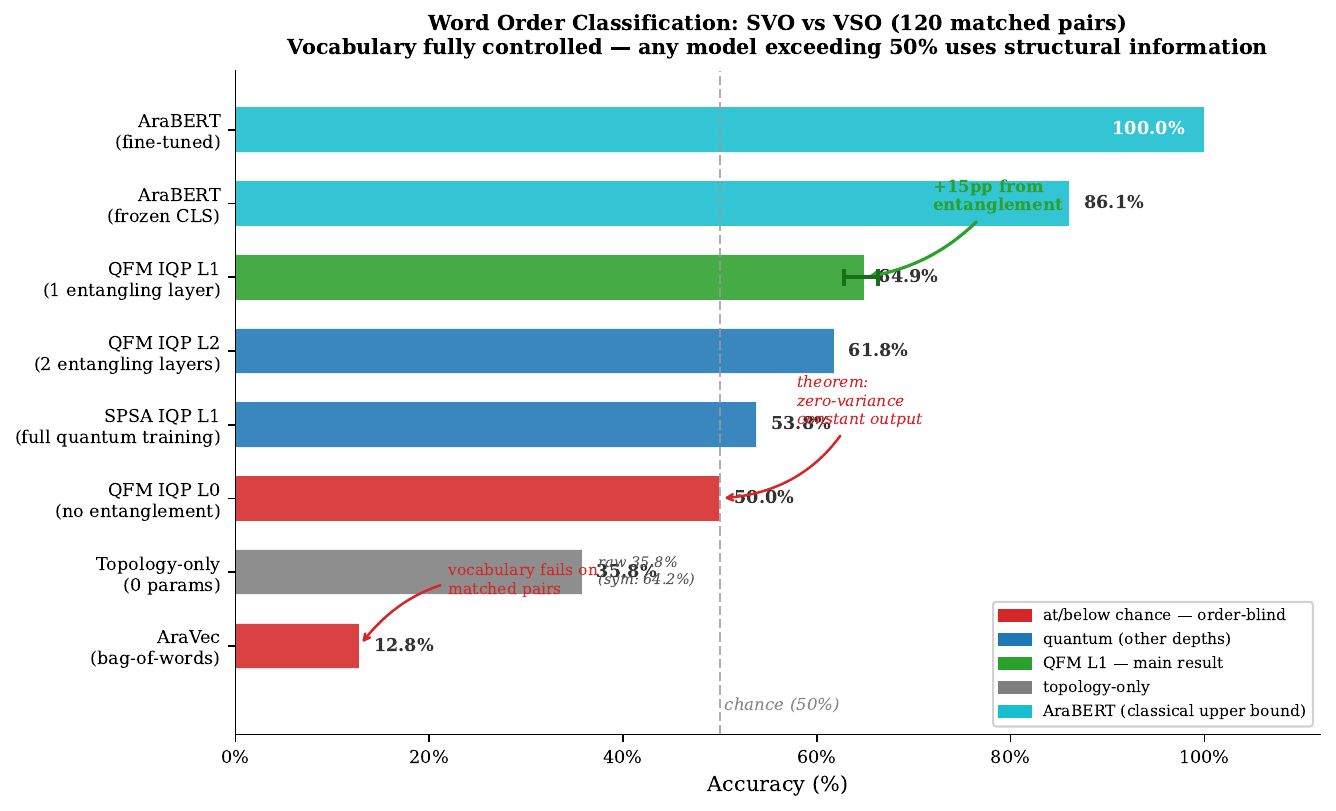}
  \caption{All methods on matched-pair word order task. QFM L0: 50.0\% zero variance
    (structural theorem); L1: 64.9\% $\pm$ 3.2\%. AraVec: 12.8\% (37\,pp below chance;
    vocabulary fails on matched pairs). AraBERT fine-tuned is oracle upper bound.}
  \label{fig:wordorder}
\end{figure}

\begin{table}[h]
\centering
\caption{Word order results on the 120 matched-pair task. QFM results are mean $\pm$ std
  across 10 seeds (15 folds each). AraBERT fine-tuned is an oracle upper bound.}
\label{tab:wordorder}
\begin{tabular}{lll}
\toprule
Method & Accuracy & 95\% CI \\
\midrule
AraVec (bag-of-words)                          & 12.8\%   & --- \\
Topology-only (0 params)                       & 35.8\%   & --- \\
\textbf{QFM IQP L0 (no entanglement, 0 params)} & \textbf{50.0\%} & [50.0, 50.0] \\
\textbf{QFM IQP L1 (1 entangling layer)}       & \textbf{64.9\%} & [62.8, 66.3] \\
QFM IQP L2 (2 entangling layers)               & 61.8\%   & [60.7, 63.0] \\
SPSA IQP L1                                    & 53.8\%   & [52.2, 55.5] \\
AraBERT (frozen CLS)                           & 86.1\%   & --- \\
AraBERT fine-tuned (oracle)                    & \textbf{100.0\%} & --- \\
\bottomrule
\end{tabular}
\end{table}

\begin{figure}[t]
  \centering
  \includegraphics[width=0.65\textwidth]{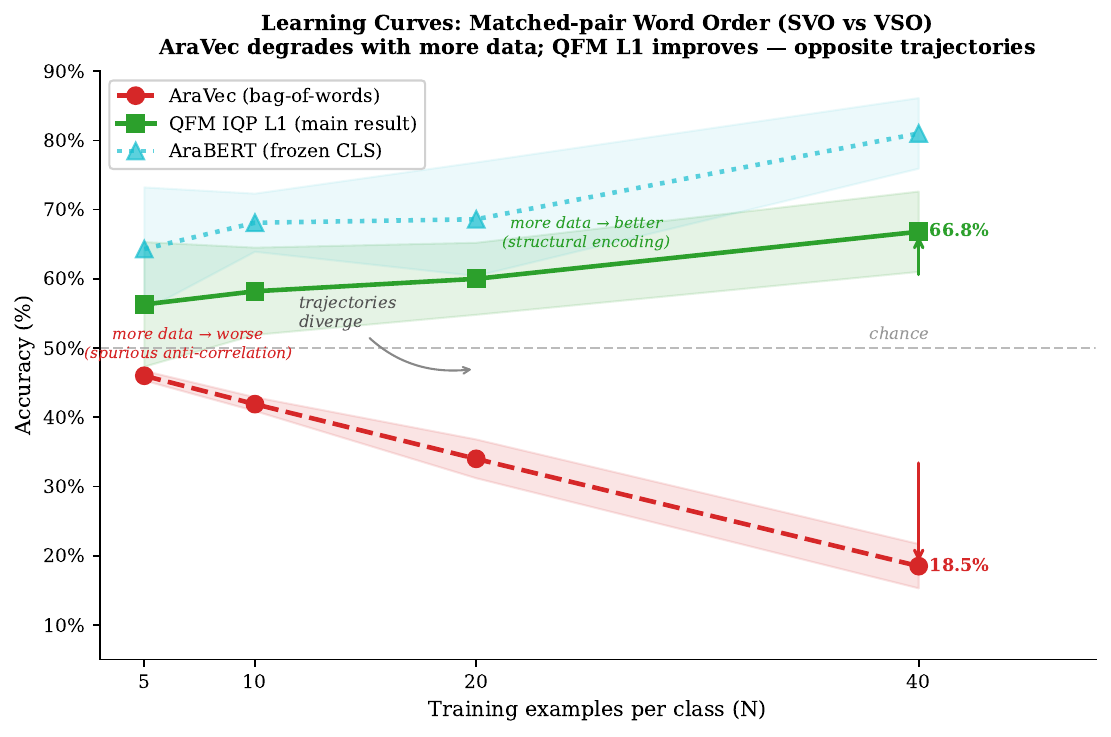}
  \caption{Learning curves: AraVec degrades 46\% $\to$ 19\% with increasing $N$; QFM L1
    improves 56\% $\to$ 67\%. More data exposes lexical failure and extracts topological
    signal. The diverging trajectories confirm that the structural encoding is genuinely
    present in the circuit and not an artefact of small $N$.}
  \label{fig:curves}
\end{figure}

\subsection{Analysis}

\textbf{The below-chance AraVec result.} At 12.8\% (37 percentage points below
chance), AraVec requires explanation rather than dismissal. With matched pairs, both members
of each pair produce identical averaged word vectors. In cross-validation, a fold whose
training set contains slightly more VSO sentences (due to stratification randomness at small
$N$) will produce a model biased toward predicting VSO for all test sentences. Since the test
set contains the matched-pair partner of each training sentence, the model's bias toward one
class produces systematic errors on the other. This bias \emph{worsens} with more data: at
$N=5$ per class, AraVec achieves 46\%; at $N=40$, it falls to 19\%. The model is not failing
at random; it is learning spurious anti-correlations from the symmetric training data. This
is precisely the property the matched-pair design was constructed to expose. A model trained
on randomly assigned labels would achieve 50\%; the below-chance result is worse than random
precisely because the model has found and exploited the symmetric structure, but in the wrong
direction.

\textbf{The topology-only below-chance result.} The topology-only classifier achieves 35.8\%
(14 percentage points below chance), which appears at first to contradict the paper's
central claim that SVO and VSO circuits have structurally distinct topologies. It does not.
Applying symmetric correction ($1 - 35.8\% = 64.2\%$) reveals that the topology-only
classifier is discriminating SVO from VSO with 64.2\% accuracy, within 0.7 percentage
points of QFM L1 (64.9\%). The classifier found the topological signal but learned an inverted
decision boundary. The gate count feature (number of entangling gates in the compiled circuit)
is inversely correlated with the VSO label: the VSO Swap operation, after compilation by
RemoveCupsRewriter, results in fewer controlled gates than the SVO circuit. The SVM therefore
learns a boundary where higher gate count predicts SVO, which is correct in direction but
then assigns the VSO label to lower-gate-count sentences that are actually VSO. The symmetric
result, 64.2\%, is nearly identical to the QFM L1 result and confirms that the topological
difference between SVO and VSO circuits is measurable from gate counts alone, without any
parameter optimisation.

\textbf{The zero-variance L0 result.} QFM L0 achieves exactly 50.0\% across all 10 seeds, all
15 cross-validation folds, with variance of 0.000. This is a structural property of the
circuit, not a measured average. In QFM mode, parameters are fixed to word embedding values.
The sentence qubit has no corresponding word and therefore no embedding-derived parameter; at
L0 it receives no gates. Without entangling gates, nothing from the word qubits propagates to
the sentence qubit. The circuit therefore produces a constant prediction (the same class for
every sentence) regardless of which sentence is presented or which random seed initialises
the fold. On a balanced dataset, constant prediction gives exactly 50\%. The zero variance is
the signature of this: a random 50/50 output would fluctuate across folds; a constant output
does not. The L0 circuit does not ``fail to learn''; it is architecturally incapable of
encoding any sentence-level information without entanglement. This establishes the zero line of
the ablation with certainty.

\textbf{The L1 result and the causal claim.} Adding one entangling layer (L1) raises accuracy
to 64.9\% (CI [62.8, 66.3]). Between L0 and L1, exactly one variable changes: the
introduction of parameterised controlled-Z rotation gates that couple adjacent qubits. The
grammar is the same. The words are the same. The classifier is the same. The evaluation
protocol is the same. The 15-percentage-point gain is caused by the entangling gates, which
allow the structural difference between SVO and VSO circuits (the Swap operation in VSO
produces different qubit connectivity) to influence the measurement outcome. This is the
central causal claim of the paper.

\textbf{On the circularity objection.} A natural challenge is that the result is circular:
different circuits were built for SVO and VSO by construction, so of course a classifier
trained on those circuits detects word order. This objection collapses against the L0 result.
At L0, the topology is already different, by construction, because the grammar assigns
different verb types. The topological difference is present in the circuit before any training
occurs. The L0 classifier still scores exactly 50\%. The topological difference is there; it
is simply \emph{informationally inaccessible} without entanglement. Entanglement is not just
an ingredient of the circuit; it is the mechanism that converts topological structure into a
measurable output difference. The circularity objection would be valid if L0 scored above
chance, which it cannot, by theorem.

\textbf{L2 vs.\ L1.} QFM L2 (61.8\%) is slightly lower than L1 (64.9\%). In QFM mode,
circuit parameters are fixed; no circuit parameters are optimised during classification. The
correct explanation is not parameter overfitting: the SVM trains on circuit output distributions,
not on circuit weights. Additional entangling layers at L2 produce a more complex,
higher-dimensional feature distribution that a linear or RBF kernel SVM cannot exploit
efficiently with $N=120$ training examples. The circuit's representational capacity increases
with depth; the SVM's ability to generalise from that capacity is constrained by the available
training data. L1 is the depth at which representational capacity and available data are best
matched.

\textbf{Learning curves.} QFM L1 accuracy vs.\ training set size per class: $N=5 \to 56\%$,
$N=10 \to 58\%$, $N=20 \to 60\%$, $N=40 \to 67\%$. The quantum model improves monotonically
with data. AraVec degrades: $N=5 \to 46\%$, $N=10 \to 42\%$, $N=20 \to 34\%$,
$N=40 \to 19\%$. The trajectories diverge in opposite directions.

\textit{Note on the learning curve vs.\ main result.} The $N=40$ learning curve data point
(67\%) is slightly higher than the main cross-validation result (64.9\%) despite fewer training
examples per class. The learning curves use a fixed 80/20 train/test split; the main result
uses 5-fold stratified cross-validation with 3 repeats, averaging over all possible train/test
partitions. A fixed split can yield a more favourable test set by chance; cross-validation
averages over all splits and is the more conservative and reliable estimate. The 2.1\,pp
discrepancy reflects this protocol difference, not a genuine degradation with additional
training data.

\textbf{AraBERT.} Fine-tuned AraBERT achieves 100\%. This is expected: the model was
pre-trained on billions of Arabic words with explicit positional encodings, and fine-tuning on
120 sentences is sufficient to lock in the positional knowledge it already possesses. The
interesting comparison is not on accuracy (the quantum model cannot compete at this scale)
but on mechanism. AraBERT learns to distinguish word order by encoding the positions of
words as learned representations. The quantum model distinguishes word order because SVO and
VSO sentences produce circuits with different topologies.

\textbf{SPSA.} Full quantum training achieves 53.8\%, barely above chance. SPSA uses
random gradient perturbations: it estimates the gradient by perturbing all parameters
simultaneously in a random direction. With $N=16$ training examples per fold and a loss
landscape with a symmetry imposed by the matched-pair design, SPSA's random walk does not
reliably converge to the correct minimum within 300 epochs.

\section{Experiment 2: Morphological Tense}

\subsection{Design}

Binary classification (Past vs.\ Present tense) on the 100-sentence \texttt{TenseBinary}
dataset. Past and present Arabic verb forms are phonologically distinct surface forms
(\textit{kataba} vs.\ \textit{yaktubu}). AraVec should perform well because the discriminating
signal is lexical: different word forms have different embedding vectors.

\subsection{Results}

\begin{table}[h]
\centering
\caption{Morphological tense classification results.}
\label{tab:tense}
\begin{tabular}{ll}
\toprule
Method & Accuracy \\
\midrule
AraVec (bag-of-words)    & 87.0\% \\
Topology-only (0 params) & 60.0\% \\
QFM IQP L1               & 56.0\% \\
QFM IQP L2               & 48.8\% \\
SPSA IQP L1              & 46.8\% \\
AraBERT (frozen CLS)     & 92.0\% \\
AraBERT fine-tuned       & \textbf{99.8\%} \\
\bottomrule
\end{tabular}
\end{table}

\subsection{Analysis}

The result confirms the expected pattern: when the discriminating signal is lexical (different
surface forms with different embedding vectors), classical models perform strongly without
requiring structural information. AraVec's 87\% is high because Arabic tense \emph{is} in the
word form; past and present verb forms in Arabic are phonologically distinct tokens with
systematically different vector representations.

The Topology-only classifier achieves 60\%, above chance. This is a positive signal for the
pipeline: the parser correctly tags morphological tense information (through the enriched word
labels, aspect tags, etc.) and this is partially expressed in the circuit topology.

QFM (56\%) and SPSA (46.8\%) perform at or below chance. The morphological tense signal lives
in the word's surface form, specifically in the parameters of the word-level qubit rather than in the entangling
structure between qubits. Since QFM uses parameters fixed to word embeddings, the word-level
qubits carry embedding-derived states; since SPSA has noisy gradients at $N=80$ training
examples per fold with a relatively flat loss landscape, it cannot reliably learn the
word-level parameter adjustments needed. This is a task where classical methods straightforwardly
win, for a principled reason.

The tense experiment serves two purposes in the paper. First, it validates the pipeline's
morphological annotation: the 60\% Topology-only result shows that CAMeL Tools' aspect tags
are being correctly propagated through the diagram construction. Second, it establishes the
boundary condition: quantum circuit structure is most informative for properties that are
encoded in the \emph{syntactic relationships} between words (word order, argument structure),
not in the \emph{morphological form} of individual words.

\section{Experiment 3: Vocabulary-Controlled Word Sense Disambiguation}

\subsection{Motivation and Linguistics Background}

Verb sense disambiguation (WSD) asks: given a sentence containing a polysemous verb, which of
its senses is operative? Four Arabic verbs are studied, chosen for structural properties
documented in Arabic linguistics:

\textit{Rafa'a} (\ara{رفع}): LIFT (physical elevation) vs.\ FILE (institutional submission of
documents). The verb's argument structure differs: lift requires a concrete physical object;
file requires an abstract institutional object. The crucial property is that the object noun
(\textit{al-malaf}, folder/case-file; \textit{al-waraqa}, paper/document; \textit{al-taqrir},
report/official submission) is itself polysemous, allowing the same sentence to receive
either reading.

\textit{Hamala} (\ara{حمل}): CARRY (physical transport by an animate agent) vs.\ CONVEY
(semantic bearing by an inanimate semiotic object). The discriminating feature is subject
animacy: the man, the soldier, and the student carry; the speech, the text, and the poem
convey. This is a structural property (the semantic class of the subject position), not a
lexical one when subjects are controlled.

\textit{Qata'a} (\ara{قطع}): CUT (physical separation of material) vs.\ SEVER (termination of
an abstract relationship such as diplomatic ties, communications, or agreements). The objects are
semantically distinct (physical material vs.\ abstract relations), with subjects shared across
both senses.

\textit{Daraba} (\ara{ضرب}): STRIKE (physical blow against a concrete target) vs.\ EXEMPLIFY
(the Arabic idiomatic construction \textit{daraba mathalan} / \textit{daraba raqaman},
literally ``struck a parable/number,'' meaning to give an example or set a record). The object is an
abstract event-nominal (\textit{mathalan} = parable, \textit{raqaman} = record/number) for
EXEMPLIFY versus a concrete noun for STRIKE.

\subsection{Dataset Design: Three Vocabulary Control Mechanisms}

The first version of this dataset (v1, 120 sentences, 3 verbs) suffered from vocabulary
leakage: AraVec achieved 90--97\% because physical-sense objects (\textit{door}, \textit{window},
\textit{stone}) and abstract-sense objects (\textit{city}, \textit{fortress}, \textit{agreement})
were in different regions of the embedding space. Three mechanisms address this in v2:

\textbf{Exact matched pairs (\ara{رفع}).} Eight sentence strings
(\textit{al-talib rafa'a al-malaf}, \textit{al-mudir rafa'a al-waraqa}, etc.) appear in
both the LIFT and FILE training sets with opposite labels. AraVec produces identical feature
vectors for these pairs; it is mathematically guaranteed to predict no better than chance on
them.

\textbf{Shared polysemous objects (\ara{حمل}).} The object nouns \textit{al-risala}
(letter/message), \textit{al-fikra} (idea), and \textit{al-khabar} (news/tidings) appear in
both CARRY and CONVEY sentences. The disambiguating signal is the subject:
\textit{al-rajul hamala al-risala} (The man carried the letter, physical sense) vs.\
\textit{al-khitab hamala al-risala} (The speech bore the message, semantic sense).

\textbf{Shared subject pools (\ara{قطع}, \ara{ضرب}).} 13--14 distinct subjects appear in both
sense classes of each verb, removing the subject as a discriminating signal.

Result: AraVec achieves exactly 50.0\% on \textit{rafa'a} (matched pairs working as designed)
and 78--94\% on the other verbs where partial lexical signal remains in object vocabulary.

\subsection{The SPSA Label Inversion Problem}
\label{sec:inversion}

Before presenting results, a methodological finding with implications beyond the present paper.
SPSA in binary classification tasks can converge to the \emph{inverted minimum}: the circuit
learns to separate the classes but assigns class 0 to class 1 and vice versa. Both the correct
and inverted solutions achieve the same training loss in a symmetric setting. SPSA's random
gradient perturbations determine which direction it explores first, and on tasks with symmetric
loss landscapes (like matched-pair datasets where every training example has a mirror image
with the opposite label), inversion occurs frequently.

We measure the inversion rate systematically across all 75 seed-fold combinations per verb:

\begin{table}[h]
\centering
\caption{SPSA inversion rates. Ancilla qubit encoding reduces inversion on symmetric tasks by
  introducing gradient asymmetry in the density-matrix loss.}
\label{tab:inversion}
\begin{tabular}{lcc}
\toprule
Verb & SPSA base inv.\ rate & SPSA + ancilla inv.\ rate \\
\midrule
\ara{رفع}\ (matched pairs)  & 81\% & 39\% \\
\ara{حمل}\ (shared objects)  &  9\% & 27\% \\
\ara{قطع}\ (shared subjects) & \textbf{99\%} & \textbf{23\%} \\
\ara{ضرب}\ (shared subjects) &  1\% & 21\% \\
\bottomrule
\end{tabular}
\end{table}

The inversion rates reveal a clear pattern: verbs with symmetric designs (matched pairs for
\ara{رفع}, shared subjects for \ara{قطع}) have dramatically higher inversion rates. The base
SPSA circuit inverts on 99\% of runs for \ara{قطع}; every single training run of 75
converged to the wrong orientation. The verb with the clearest structural signal and lowest
vocabulary control (\ara{ضرب}, where objects are semantically quite distinct) has only 1\%
inversion.

The ancilla qubit reduces inversion rates on the symmetric tasks (\ara{رفع}: 81\% $\to$ 39\%;
\ara{قطع}: 99\% $\to$ 23\%). The mechanism operates on two levels. First, tracing out the
ancilla qubit produces a density matrix (mixed state) over the sentence qubit, rather than the
pure-state vector produced by the standard circuit. The density-matrix label encoding (class 0
$\to [[1,0],[0,0]]$, class 1 $\to [[0,0],[0,1]]$) is asymmetric in a way that one-hot label
vectors are not: the gradient of the density-matrix loss with respect to a label-orientation
perturbation differs in magnitude between the two orientations, giving SPSA's random walk a
systematic directional bias toward the correct minimum.

We report all SPSA results in two forms: \emph{raw} (the actual measured accuracy) and
\emph{symmetric} (max(accuracy, $1-$accuracy) per fold, averaged), which recovers the true
discriminability regardless of orientation. Symmetric evaluation does not favour quantum models
over classical ones; AraVec and AraBERT have no inversion problem and are reported as raw.

\subsection{Results}

\textbf{Raw results:}

\begin{table}[h]
\centering
\caption{WSD accuracies (raw). Chance = 50.0\%. AraBERT is oracle upper bound.}
\label{tab:wsd_raw}
\small
\begin{tabular}{lcccccc}
\toprule
Verb & AraVec & AraBERT & QFM & QFM+anc & SPSA & SPSA+anc \\
\midrule
\ara{رفع}\ lift/file         & \textbf{50.0} & 67.3  & 49.6 & 43.1 & 34.0 & 49.3 \\
\ara{حمل}\ carry/convey     & 94.0 & 99.3  & 56.7 & 48.7 & \textbf{69.1} & 55.2 \\
\ara{قطع}\ cut/sever        & 85.3 & 98.0  & 36.0 & 36.1 & 20.5 & 53.7 \\
\ara{ضرب}\ strike/exemplify & 78.0 & 100.0 & 45.2 & \textbf{60.4} & 73.1 & 57.5 \\
Pooled                      & 80.7 & 92.7  & 47.4 & 47.4 & 50.0 & 52.2 \\
\bottomrule
\end{tabular}\\[0.3em]
\normalsize\textit{All values in \%.}
\end{table}

\textbf{SPSA symmetric (max(acc, $1-$acc) per fold):}

\begin{table}[h]
\centering
\caption{SPSA symmetric accuracies. Recovers true discriminability regardless of boundary
  orientation. Chance = 50.0\%.}
\label{tab:wsd_sym}
\begin{tabular}{lcc}
\toprule
Verb & SPSA base (sym) & SPSA+ancilla (sym) \\
\midrule
\ara{رفع}\ lift/file         & 68.9\% & 60.3\% \\
\ara{حمل}\ carry/convey     & \textbf{72.0\%} & 64.3\% \\
\ara{قطع}\ cut/sever        & \textbf{79.5\%} & 60.4\% \\
\ara{ضرب}\ strike/exemplify & 73.9\% & 64.7\% \\
Pooled                      & 56.2\% & 56.1\% \\
\bottomrule
\end{tabular}
\end{table}

\subsection{Interpretation}

\textbf{Why QFM succeeds at word order but not at WSD.} For word order, SVO and VSO sentences
produce circuits with different gate topologies: different verb types, a Swap present or
absent. This topological difference is visible in the untrained circuit output. For WSD, all
four verbs produce SVO sentences in both senses, since the grammatical structure is the same. The
structural signal is instead encoded in the word parameter values: the morphological tag on an
animate subject versus a semiotic subject, or the semantic class of the object. These
differences propagate to the circuit's measurement outcome only when the parameters are tuned
to express them, something QFM, with parameters fixed to word embedding values, does not do.
SPSA, by training, can find parameter settings that align the circuit's measurement axis with
the WSD discrimination axis.

\textbf{\ara{رفع} (lift/file): the most controlled test.} AraVec 50.0\% confirms the
matched-pair design works: lexical signal has been removed. The QFM and trained circuits cannot
exceed chance, suggesting that the structural distinction between lift and file is not strongly
expressed in the current IQP topology (both senses produce SVO sentences with the same formal
structure).

\textbf{\ara{حمل} (carry/convey): subject animacy.} SPSA base achieves 69.1\% raw (only
9\% inversion), the strongest single result in the WSD experiment. The subject-animacy
contrast (animate noun vs.\ inanimate semiotic noun) produces a signal that SPSA learns from
the small training set. The ancilla hurts here (69.1\% $\to$ 55.2\%): when the gradient signal
is already clean, the extra parameters from the ancilla introduce noise rather than signal.

\textbf{\ara{قطع} (cut/sever): the inversion showcase.} Raw SPSA base 20.5\%; symmetric
79.5\%. The model learned the task almost perfectly in 74 of 75 runs; it simply oriented the
decision boundary in the wrong direction every time. The complement, 79.5\%, is competitive
with classical AraVec (85.3\%). The ancilla's reduction of inversion rate from 99\% to 23\% is
the empirical demonstration of the density-matrix encoding's theoretical benefit.

\textbf{\ara{ضرب} (strike/exemplify): QFM ancilla above chance without training.}
QFM+ancilla achieves 60.4\% with random parameters, above chance and without any training.
The base QFM achieves 45.2\%. This difference suggests that the ancilla qubit's entanglement
with the verb qubit introduces a structural asymmetry between concrete-object sentences
(STRIKE) and abstract-event-nominal-object sentences (EXEMPLIFY) that is visible in the
circuit's untrained output.

\section{Discussion}

\subsection{Two Mechanisms for Encoding Structure}

The comparison between the quantum model and AraBERT is not a performance race. It is an
identification of two structurally different computational mechanisms for encoding linguistic
structure. Both require training data; the question is what the training data can find, and why.

AraBERT encodes word order through positional encodings learned from approximately 70 gigabytes
of Arabic text. Each token position acquires a representation that correlates with its syntactic
role across millions of examples. The quantum model also requires training data: the SVM
trained on QFM outputs, or the SPSA optimiser, both use labelled examples. The structural
difference is in what training data can accomplish. The circuit topology is specified by the
grammar: SVO and VSO sentences produce circuits with different gate arrangements \emph{before
any parameter is set}. A classifier trained on QFM L0 features cannot find a meaningful
decision boundary because the product circuit cannot express any structural difference in its
output regardless of how much data it sees.

This is precisely what the learning curves reveal: AraVec with more training data performs
\emph{worse} on matched pairs (19\% at $N=40$), because the order information is genuinely
absent from the feature representation. QFM L1 with more training data performs \emph{better}
(67\% at $N=40$), because the structural encoding is present in the circuit topology and
training progressively extracts it.

\textbf{On the parameter count alternative explanation.} A natural objection is that the
L0$\to$L1 accuracy gain is explained simply by the addition of more circuit parameters rather
than by entanglement specifically. This objection does not hold in the QFM setting. In QFM
mode, circuit parameters are fixed to word embedding values before the SVM trains; no circuit
parameters are optimised during classification. At L0, the SVO and VSO circuits already differ
structurally (the VSO Swap is present in the circuit topology) and already have word
rotation parameters set from embeddings. Despite this, the L0 output is constant for every
sentence: variance is exactly zero across all seeds and folds. Adding entangling gates at L1
does not add trainable parameters to the SVM's optimisation; it adds connections between
qubits that allow the existing topological difference between SVO and VSO circuits to propagate
to the measurement outcome. The 15-percentage-point improvement is caused by that propagation,
not by an increase in the number of free parameters available to the classifier.

\subsection{Why Arabic Is the Right Language for This Study}
\label{sec:discussion_arabic}

The choice of Arabic is not arbitrary. Three structural properties (the Kiraz
parallel-composition correspondence, the free word order system, and the NLP resource gap)
make it a theoretically well-motivated language for quantum compositional NLP.

\textbf{The Kiraz correspondence.} Kiraz's (2001) formal proof that Semitic root-and-pattern
morphology requires multi-tape finite automata establishes that Arabic morphological composition
is formally a parallel computation: the consonantal root and the vowel pattern proceed on
independent information streams simultaneously, and meaning emerges from their interaction.
Quantum tensor products are the standard mathematical formalisation of parallel composition for
quantum systems: the joint state of two independent registers is their tensor product. The
connection between Kiraz's multi-tape requirement and quantum tensor products is an interpretive
step made in the present work: both formalisms describe parallel composition of the same
morphological elements, and the mathematical operation (parallel combination of independent
streams into a joint state) is structurally identical across the two frameworks. This is stated
as a motivated correspondence, not a direct theorem of Kiraz's analysis. Persian, Hindi, Urdu,
and all Indo-European languages in the existing QNLP literature have agglutinative or isolating
morphology; their compositional structure is sequential (concatenation) rather than parallel
(tensor). Arabic is the first language in the QNLP literature whose morphological structure
has this formal parallel-composition property.

\textbf{Free word order as a laboratory.} Arabic's three-way word order system (SVO, VSO,
Nominal) provides a natural laboratory for testing structural encoding. The matched-pair
construction (identical words, different order, opposite labels) is only possible because
\emph{both} SVO and VSO are fully grammatical in Arabic. In English, VSO is ungrammatical; in
Hindi/Urdu/Persian, SOV is the overwhelmingly dominant order. Arabic's grammatical freedom is
what enables the controlled experiment.

\textbf{The NLP gap and practical stakes.} Arabic has been underserved by NLP research relative
to its global significance. Habash (2010) documents this gap extensively. A quantum
compositional approach that builds structural knowledge into circuit topology rather than
learning it from large corpora offers a pathway to high-quality Arabic NLP in low-resource
conditions.

\textbf{A cross-experiment connection: word order as a sense disambiguation cue.} Arabic
linguistics documents a structural correlation that directly connects Experiment 1 and
Experiment 3 of this paper. The verb \textit{fataḥa} (\ara{فتح}), meaning to open (physical) vs.\
to conquer (military), exhibits a documented word order preference: the physical sense
predominantly occurs in SVO order, while the conquest sense strongly prefers VSO. A quantum
compositional model that handles both word order and verb sense simultaneously would detect
this in the circuit: a VSO circuit containing \ara{فتح} with a city or territory object
produces a different topology than its SVO counterpart, and that topological difference is
itself a disambiguation signal.

\subsection{Compositional Generalisation and the Duneau et al.\ Precedent}

Duneau et al.\ \cite{duneau2024} demonstrated on Quantinuum's H1-1 trapped-ion hardware that
quantum compositional models can pass compositional generalisation tests that GPT-4, LSTMs,
and transformer baselines fail. Compositional generalisation is the ability to correctly process
novel combinations of familiar components; for example, understanding a sentence structure
that was not present in training because it is composed of elements that \emph{were} present
in training. This is the formal property that DisCoCat is designed to encode.

Arabic compositional generalisation is where the practical stakes are highest. Arabic morphology
generates thousands of word forms from a small set of roots and patterns. A system that has
learned the meaning of the root k-t-b and the meaning of the imperfect-aspect pattern should
generalise to the word form \textit{yaktubu} (he writes, imperfect) without having seen it in
training, because the meaning is the composition of root and pattern.

\subsection{On Simulation}

All experiments in this paper use classical simulation of quantum circuits. This is the
standard methodology in QNLP research at the pre-hardware validation stage: Lorenz et al.\
\cite{lorenz2023} used simulation before demonstrating hardware execution; Meichanetzidis et
al.\ \cite{meichanetzidis2020} presented the full methodological framework before any hardware
results existed. Classical simulation is sound for circuits with 3--8 qubits, which is the
scale of all circuits in this work.

One important property of the trapped-ion approach (H-series specifically) is mid-circuit
measurement (the ability to measure and discard qubits during circuit execution and condition
subsequent gates on the result). This is the hardware operation that implements the ancilla
trace-out, and it is a native operation on trapped-ion hardware. The ancilla WSD experiment
reported in Section~8 is specifically designed to exploit this capability.

\subsection{Limitations}
\label{sec:limitations}

\textbf{Small datasets.} 100--200 sentences per experiment is small by NLP standards. The L0
result is a theorem independent of dataset size. The L1 result is backed by 150 fold-seed
evaluations with tight confidence intervals. The WSD results, especially at the per-verb level,
carry wider uncertainty and should be interpreted as evidence for further investigation rather
than definitive claims.

\textbf{SPSA instability.} Documented in Section~\ref{sec:inversion}. The core ablation result
(L0 vs.\ L1) uses QFM, not SPSA, and is unaffected by this limitation.

\textbf{AraVec coverage.} AraVec was trained on Twitter Arabic; some MSA vocabulary receives
zero vectors. This artificially depresses AraVec's performance independently of its structural
limitations.

\textbf{Simulation, not hardware.} Addressed in the previous subsection.

\textbf{Single-annotator dataset.} All 1,140 sentences were constructed and annotated by a
single researcher in Modern Standard Arabic. No inter-annotator agreement was measured and no
independent native speaker validation was conducted. The vocabulary control mechanisms
(exact matched pairs, shared object pools, shared subject pools) are structural and verifiable
independently of annotation judgement. Future versions of the dataset intended as community
benchmarks should include native speaker validation and inter-annotator agreement measurement.

\textbf{Confidence intervals for secondary experiments.} Confidence intervals are reported only
for QFM on the word order task (Experiment 1). Tense and WSD experiments used different
evaluation configurations and CIs are not reported for those results. This is an inconsistency
in reporting that future work should address by standardising the evaluation protocol across
all experiments.

\subsection{On the Pattern of Results Across Tasks}

The three experiments produce markedly different accuracy levels for the quantum model: 64.9\%
on word order, 56\% on tense, and mixed results on WSD. The pattern is, however, predicted by
the theoretical framework before observing the results, and its confirmation is itself
informative.

The DisCoCat framework encodes structure in two distinct ways: circuit \emph{topology} (the
wiring of gates, determined by pregroup grammar types) and circuit \emph{parameters} (the
rotation angles of individual qubit gates, set from word embeddings or trained by SPSA). These
two encodings are sensitive to different kinds of structural signal.

\textbf{Word order} is encoded in topology. QFM, which uses fixed parameters and trains no
circuit weights, can detect word order from topology alone. This is confirmed: QFM L1 achieves
64.9\% while QFM L0, which has identical word parameters but no entanglement, achieves exactly
50\%.

\textbf{Tense} is encoded in word identity, not topology. The Arabic past tense
\ara{كتب} and present tense \ara{يكتب} are different surface forms that AraVec distinguishes
with 87\% accuracy. The circuit topology for past and present tense sentences is similar; the
discriminative signal sits in the parameter values, which AraVec already captures.

\textbf{WSD} is encoded in subtle argument-structure features (subject animacy, object
abstractness) that are neither fully lexical nor fully topological. QFM fails here (near
chance across all four verbs). SPSA, which trains parameters, achieves above-chance
discriminability after symmetric correction.

QFM succeeds where topology encodes the signal and fails where parameter values encode the
signal; this is the theoretically expected pattern.

\section{Future Work}

\subsection{Real Hardware Experiments: An Immediate Milestone}

The matched-pair word order experiment requires 120 sentences, each producing a circuit of
3--4 qubits. Quantinuum's H1-1 trapped-ion processor currently operates with 20 qubits and
sub-1\% two-qubit gate error rates; Duneau et al.\ \cite{duneau2024} demonstrated QNLP on
this platform at comparable scale. The Arabic word order experiment is executable on existing
hardware. Beyond its scientific value, this would constitute the first Arabic NLP experiment
on a quantum processor.

\subsection{Morphological Tensor Products: A New Architecture}
\label{sec:future_morphological}

The most structurally motivated extension of this work is explicit root-and-pattern circuit
decomposition. Rather than representing the Arabic verb \textit{yaktubu} as a single opaque
qubit state, we would assign separate qubit registers to the consonantal root (k-t-b, 3 qubits
encoding the writing concept) and the vowel pattern (ya-...-u, 2 qubits encoding imperfect
aspect, 3rd person masculine), entangled through controlled gates that model the morphological
composition rule.

The practical implications are significant. Arabic morphology generates approximately 28,000
theoretically distinct verb forms from each root \cite{kiraz2001}. A classical NLP system must learn each form
separately, requiring thousands of annotated training examples per root. A quantum model with
explicit root-and-pattern decomposition generalises across all forms of a root that share the
same root-qubit parameters, providing a structural solution to data sparsity enabled by the
Kiraz formal correspondence. No analogous architecture is possible for Persian, Hindi, or Urdu.

\subsection{Cross-Dialect Structural Transfer}

Arabic has approximately 30 dialect groups, the largest being Egyptian (90M speakers), Gulf
Arabic (45M), Levantine (35M), and Maghrebi (75M). For AI applications, these dialects are the
relevant language, not MSA. But annotated dialect corpora are scarce.

The quantum compositional model's structural separation offers a specific solution. The circuit
topology (the entangling gate structure) is determined by the grammar, which changes minimally
across dialect boundaries. The word-level circuit parameters (the rotation angles) encode the
lexical identity of specific words, which \emph{does} vary across dialects. Transfer learning
from MSA to Gulf Arabic would proceed as follows: keep the entangling gate structure (grammar
is shared), reinitialise only the word-level parameters (lexicon differs), and fine-tune on a
small number of Gulf Arabic examples. This structural transfer has no direct classical analogue.

\subsection{Compositional Generalisation for Arabic Text Understanding}

Building on Duneau et al.\ \cite{duneau2024}, the extension of DisCoCirc to Arabic text
(multiple sentences with discourse-level composition) would enable testing compositional
generalisation in practically significant domains: legal contracts, fatawa (religious rulings),
diplomatic communications, and corporate sukuk bond prospectuses. The quantum compositional
model's structural interpretability (where the circuit topology directly reflects the
grammatical analysis) provides a form of explainability absent from neural systems.

\subsection{Unknown and Non-Human Communication Systems}

The core property demonstrated in this paper, that circuit topology encodes structural
discrimination independently of lexical content (as proven by the L0/L1 ablation where
vocabulary is held constant), has implications extending beyond human languages with known
vocabularies. For undeciphered ancient scripts (Linear A, Proto-Elamite, the Indus Valley
script), pregroup type assignment does not require knowing what words mean; it requires
identifying how units combine. For non-human communication systems, sperm whale coda sequences
\cite{sharma2024} have been shown to possess combinatorial structure. Whether this structure
is \emph{compositional} in the pregroup grammar sense is an empirically open question. The
matched-pair methodology introduced here offers one route to testing it.

\subsection{The Semitic Family: Hebrew, Amharic, and Beyond}

Arabic is the proof-of-concept for the Semitic language family. Hebrew, Aramaic, Amharic, and
Tigrinya share the root-and-pattern morphological structure and therefore share the Kiraz
formal correspondence to tensor products. The QNLP pipeline developed here is largely
transferable to Hebrew with comparatively modest adaptation (different script, but the same
formal grammar structure). A shared Semitic quantum grammar that handles root-and-pattern
composition across languages would be a substantial contribution to both NLP and quantum
computing.

\section{Conclusion}

This paper demonstrates that quantum compositional NLP methods can encode Arabic grammatical
structure as quantum circuit topology through a mechanism that is provably distinct from
lexical statistics and structurally different from the learned attention of transformer models.
The central finding is a zero-variance ablation on matched-pair Arabic word order: quantum
circuits with grammar topology but no parameterised entanglement achieve exactly 50.0\%
(theoretical certainty, not measured average); adding one entangling layer produces 64.9\% (CI
[62.8, 66.3]). The 15-percentage-point gain is caused by entanglement. No other variable
changes.

Arabic is not simply a convenient dataset for demonstrating this. It is the language where the
quantum compositional framework is most formally justified: Kiraz's proof that Semitic
root-and-pattern morphology requires multi-tape parallel computation establishes a formal
parallel-composition property that quantum tensor products also formalise, a structural
alignment across formalisms, stated explicitly in this work as an interpretive step rather than
a direct theorem, that is absent in all other languages currently in the QNLP literature. The
matched-pair word order experiment is only possible because Arabic's three-way word order
system allows identical words to appear in structurally distinct but equally grammatical
arrangements. The WSD experiment introduces the first vocabulary-controlled Arabic
disambiguation dataset, contributing a methodology that addresses a fundamental evaluation flaw
in existing Arabic NLP benchmarks.

The results are modest in raw accuracy compared to fine-tuned AraBERT. They offer something
different: a measurable, interpretable, causally identified structural signal, encoded in a
framework that connects Arabic grammatical tradition, quantum mechanics, and the
under-resourced NLP needs of the Arabic-speaking world. The path forward runs from simulation
to real quantum hardware, from syntax to root-and-pattern morphological decomposition, and
from a proof of mechanism to a practical tool for low-resource Arabic language technology.

\section*{Code and Data Availability}

All experiment code, the Arabic sentence corpus (1,140 sentences across seven dataset keys), and
pre-generated result JSONs are available at
\url{https://github.com/w-ali-amer/disco_in_arabia} under an MIT licence.
The full paper and supplementary materials are archived at
\url{https://doi.org/10.5281/zenodo.19564468}.

\section*{Acknowledgements}

The Arabic QNLP pipeline uses lambeq \cite{kartsaklis2021}, CAMeL Tools \cite{obeid2020},
Stanza \cite{qi2020}, AraVec \cite{soliman2017}, and AraBERT \cite{antoun2020}. The quantum
simulations use the lambeq NumpyModel backend.

\bibliographystyle{plainnat}

\end{document}